\documentclass[final,5p,times,twocolumn]{elsarticle}

\usepackage{amssymb}
\usepackage[numbers]{natbib}
\usepackage{subfigure}
\usepackage{bbding}
\usepackage{amsmath}
\usepackage{multirow}
\usepackage{color, soul}
\soulregister\cite7
\soulregister\ref7

\journal{Computer Methods and Programs in Biomedicine}

\begin{document}

\begin{frontmatter}



\title{HIVE-Net: Centerline-Aware  HIerarchical View-Ensemble Convolutional  Network  for Mitochondria Segmentation in EM Images}


\author[label1]{Zhimin Yuan}
\author[label2]{Xiaofen Ma}
\author[label1]{Jiajin Yi}
\author[label1]{Zhengrong Luo}

\author[label1,label3,label4]{Jialin Peng}
\ead{2004pjl@163.com}

\address[label1]{College of Computer Science and Technology, Huaqiao University, Xiamen 361021, China}
\address[label2]{Department of Medical Imaging, Guangdong Second Provincial General Hospital, Guangzhou, 510317,  China }
\address[label3]{Xiamen Key Laboratory of Computer Vision and Pattern Recognition, Huaqiao University, Xiamen 361021, China}
\fntext[label4]{Corresponding author.}

\tnotetext[1]{This work was supported in part by by National Natural Science Foundation of China (No.11771160), and  Science and Technology Project of Fujian Province (No. 2019H0016).}

\begin{abstract}
\textit{Background and objective}: With the advancement of electron microscopy (EM) imaging technology,
neuroscientists can investigate the function of various intracellular organelles, e.g, mitochondria, at
nano-scale. Semantic segmentation of electron microscopy (EM) is an essential step to efficiently
obtain reliable morphological statistics. Despite the great success achieved using deep convolutional
neural networks (CNNs), they still produce coarse segmentations with lots of discontinuities and false
positives for mitochondria segmentation.

\textit{Methods}: In this study, we introduce a centerline-aware multitask network by utilizing centerline
as an intrinsic shape cue of mitochondria to regularize the segmentation. Since the application of 3D
CNNs on large medical volumes is usually hindered by their substantial computational cost and storage
overhead, we introduce a novel hierarchical view-ensemble convolution (HVEC), a simple alternative
of 3D convolution to learn 3D spatial contexts using more efficient 2D convolutions. The HVEC
enables both decomposing and sharing multi-view information, leading to increased learning capacity.

\textit{Results}: Extensive validation results on two challenging benchmarks show that, the proposed
method performs favorably against the state-of-the-art methods in accuracy and visual quality but
with a greatly reduced model size. Moreover, the proposed model also shows significantly improved
generalization ability, especially  when training with quite limited amount of training data. Detailed sensitivity
analysis and ablation study have also been conducted, which show the robustness of the proposed
model and effectiveness of the proposed modules.

\textit{Conclusions}: The experiments highlighted that the proposed architecture enables both simplicity
and efficiency leading to increased capacity of learning spatial contexts. Moreover, incorporating
shape cues such as centerline information is a promising approach to improve the performance of
mitochondria segmentation.

\end{abstract}

%

\begin{keyword}Electron microscopy \sep
Image segmentation \sep
Multi-task learning \sep
Centerline Detection



\end{keyword}

\end{frontmatter}


\section{Introduction}
\begin{figure}[t]
\begin{minipage}[b]{1.0\linewidth}
  \centering
  \centerline{\includegraphics[scale=0.34]{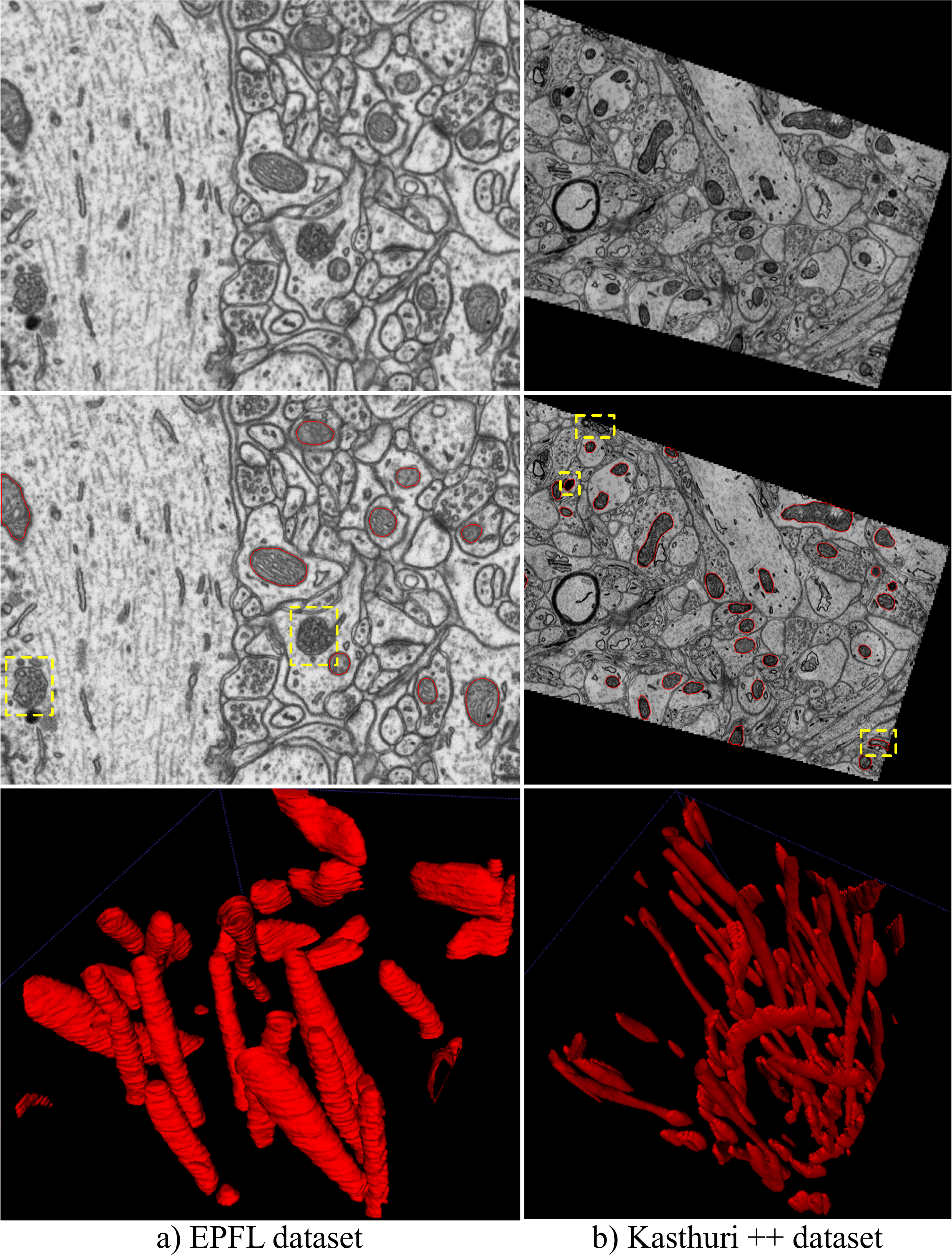}}
\end{minipage}
\caption{Illustration of challenging examples for mitochondria segmentation. First row: EM images; Second row: corresponding segmentation ground-truth (red contours); Third row: 3D visualization of ground truth. Some  error-prone locations are highlighted with dotted boxes.}
\label{fig:show_img}
\end{figure}

Mitochondria, as the powerhouse of cell, are essential sub-cellular organelles in cell's life cycle. An increasing number of studies have shown that the shape and distribution of mitochondria are closely related to neurodegenerative diseases \cite{cho2010mitochondrial} \cite{nunnari2012mitochondria}. Nowadays, with the advancement of electron microscopy (EM) imaging technology, neuroscientists can investigate the function of various intracellular organelles under high-resolution EM images at nano-scale.  Accurate delineation of mitochondria in EM images is the prerequisite step for the quantitative analysis of its morphology and distributions. However, manual delineation of such high-resolution data is time-consuming, tedious, subjective and also  has limited reproducibility \cite{perez2014workflow}.  Most importantly, biomedical images must be annotated by experts which are more costly than annotating natural images. Consequently, fully automated mitochondria segmentation algorithms with sufficient accuracy are valuable to help neurologist analyze EM images. Unfortunately, in virtue of the irregular shape variance, shift size, fuzzy boundaries of mitochondria and its complex background in EM images, as shown in Fig. \ref{fig:show_img}, mitochondria segmentation has proven to be a challenging task.

A variety of  studies have introduced kinds of specially-designed hand-crafted features \cite{peng2019mitochondria,smith2009fast,kumar2010radon,seyedhosseini2013segmentation, cetina2018multi} for mitochondria segmentation, which are usually combined with classical machine learning algorithms \cite{peng2019mitochondria,lucchi2011supervoxel,seyedhosseini2013segmentation,lucchi2013learning}, e.g., random forest \cite{peng2019mitochondria,seyedhosseini2013segmentation}, AdaBoost \cite{cetina2018multi}, conditional random fields (CRF)  \cite{lucchi2013learning,lucchi2011supervoxel}. Despite the sound performance,  approaches in this class have shown limited ability to address  the substantial increase  of data scale, due to the limited representability of hand-crafted features and small capacity of classical shallow machine learning models.

Recently, fully convolutional neural network (FCN), has  gained remarkable  performance for biomedical image segmentation \cite{ronneberger2015u,cciccek20163d,huo2018splenomegaly,zhao2018deep}. Generally,  building deeper and wider networks, also allowing a
large input size, is crucial to achieve higher accuracy for the  challenging segmentation
task. A strong baseline network for medical image segmentation is the U-Net \cite{ronneberger2015u,cciccek20163d}, which is a typical 2D FCN using encoder-decoder architecture and skip connections.
For EM segmentation, existing studies \cite{casser2018fast,xiao2018automatic,cheng2017volume,funke2018large,zeng2017deepem3d} are mainly based on the U-Net-like architecture and its 3D variants 3D U-Net \cite{cciccek20163d} and V-Net \cite{milletari2016v}. Given a volume image,  2D networks process it slice by slice. Despite its computational efficiency, the 2D segmentation network usually cannot achieve competitive segmentation performance, because the inter-slice information is completely omitted.  In contrast, networks using 3D convolutions have shown state-of-the-art results for medical volume segmentation by taking advantage of full spatial contexts. One limitation of the 3D FCN is the significantly increased number of learning parameters with 3D convolution kernel, which inevitably brings  high computational
cost. Due to GPU memory restrictions, 3D FCNs usually adopt small-scale sliding
volumes to process the original large volume.  The high GPU memory consumption of 3D FCNs also limits the network depth  compared to that of 2D FCNs.    With large parameter space, most deep-learning based algorithms, especially deep 3D FCN models, require a vast amount of manually annotated training data which is hard to collect in  clinical usage. Recently, Xie et al.\cite{xie2020sesv} proposed to predict the segmentation errors produced by an existing model and then correct them with a verification network.

Besides segmentation accuracy, computation complexity is another important consideration for model design. Since huge scale of parameters and high computational cost will inevitably make a model prohibitive on devices with limited GPU memory or application on dataset
of limited size. A typical solution for light-weight architecture design is to explore low rank factorization of convolutional kernels \cite{chollet2017xception,szegedy2016rethinking,xie2017aggregated}. For EM image segmentation,  Cheng \textit{et al.} \cite{cheng2017volume} approximated a 3D convolution  with three successive rank-1 (1D) kernels, which can greatly reduce the model parameters. However, the 1D convolutions have  limited ability to capture crucial 3D spatial context. In fact, incorporating full spatial context and multi-scale
features are critical, especially for addressing challenging tasks using light-weight models,  since EM images are cluttered with structures exhibiting similar intensities and textures, and strong gradients do not necessarily indicate the semantic boundaries of mitochondria. Therefore,  devising a  model not only effective to capture discriminative context but also efficient under strict memory and computational budget constraints  is our focus.

\begin{figure*}[t]
\begin{minipage}[b]{1.0\linewidth}
  \centerline{\includegraphics[width=0.75\textwidth]{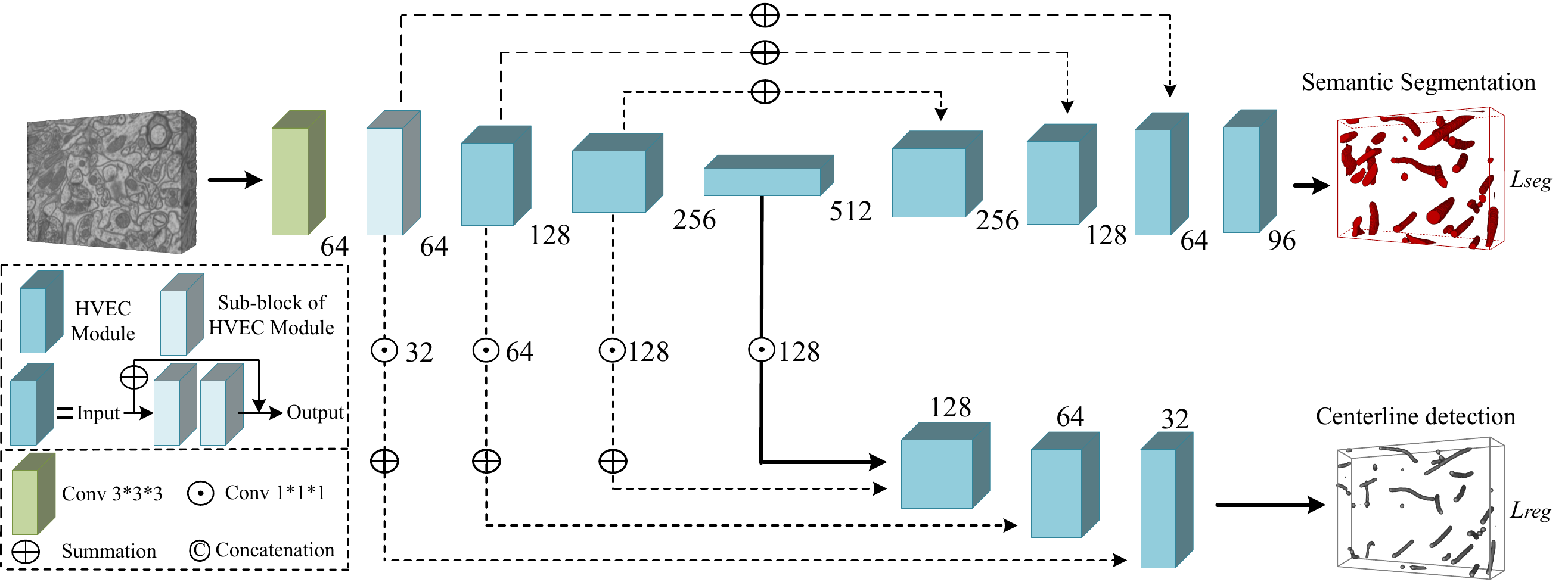}}
 \medskip
\end{minipage}
\caption{Overall architecture of our  HIVE-Net, which integrates two closely related tasks, for 3D segmentation of mitochondria. The centerline detection branch is utilized  to encode intrinsic shape knowledge, and suppress discontinuities and  false  positives in the segmentation. A novel HVEC module involving only 2D convolutions,  is the basic building blocks, which  are represented by colored cubes with numbers below being the number of feature channels.
Each HVEC module consists of two HVECs as sub-blocks with short connection cross them for residual learning.   }
\label{fig:network_img}
\end{figure*}

To address the shortcomings mentioned above, we propose a multi-task pseudo-3D network, named \textit{HIVE-Net}, which  can not only utilize intrinsic shape cues to regularize the semantic segmentation task, but also exploit 3D spatial contexts using only 2D
convolutions. Specifically, we integrate two closely related tasks, i.e. segmentation and centerline detection into a single network. The objective is to take account of the intrinsic shape cue of mitochondria represented by centerline to help improve the generalization performance and robustness of the segmentation model, especially when only scarce annotated training samples are available.

\begin{figure}[ht]
\begin{minipage}[b]{1.0\linewidth}
  \centering
  \centerline{\includegraphics[width=1\textwidth]{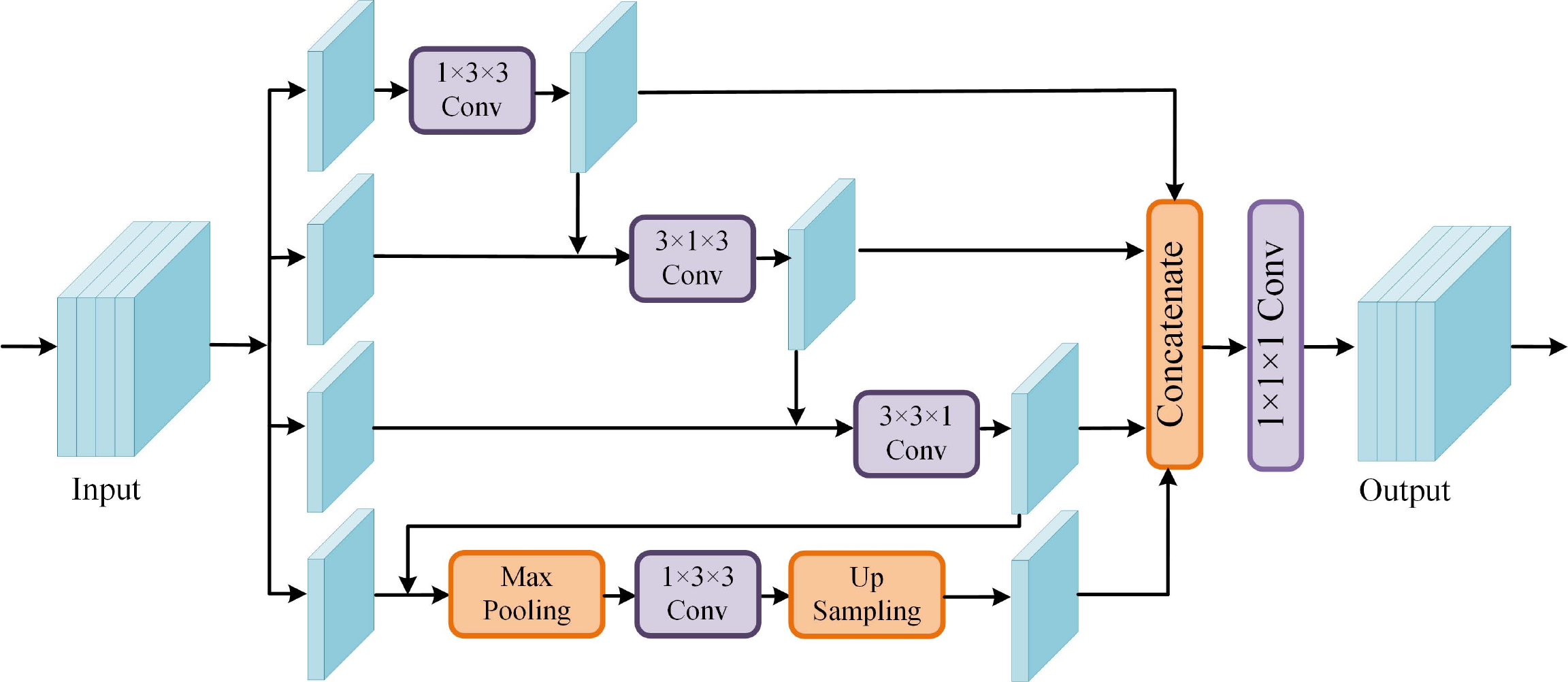}}
\end{minipage}
\caption{The multi-branch architecture of our \textit{hierarchical view-ensemble convolution} (HVEC). In order to extract multi-view multi-scale contextual features, we perform  in parallel three different rank-2 (2D) spatial convolutions, corresponding to operating on the three orthogonal views of a 3D volume,   on different subgroups of feature channels with hierarchical connections.  The fourth  branch is to extract multi-scale contextual features on one focal view. }
\label{fig:hvec_module}
\end{figure}

\begin{figure}[t]
\begin{minipage}[t]{1.0\linewidth}
  \centering
  \centerline{\includegraphics[scale=0.3]{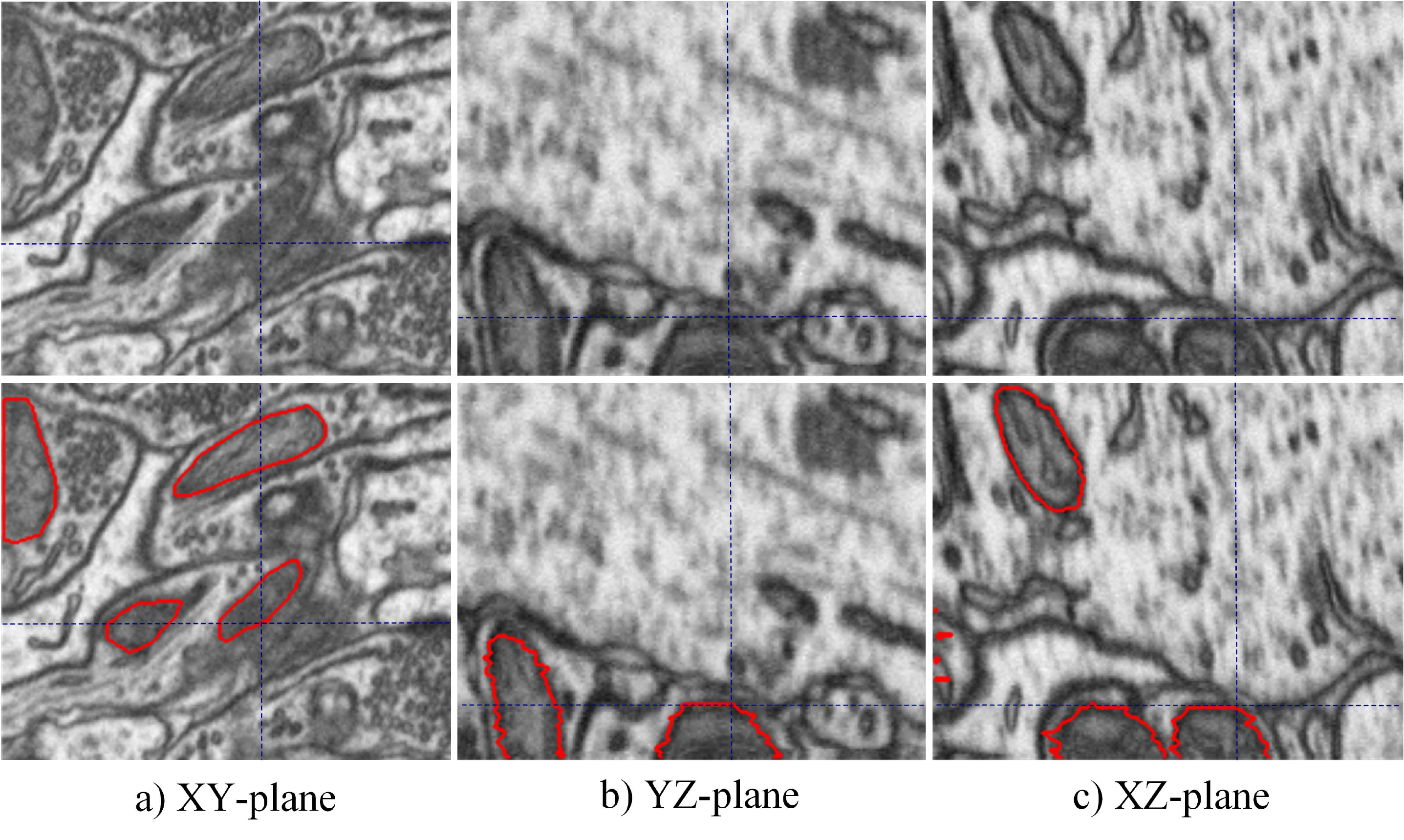}}
\end{minipage}
\caption{Illustration of the multi-view property of a 3D volume. A volumetric  image can be represented in its three dimensions using
multi-planar representation. The mitochondria with blurred boundary in the XY-plane, which can be clearly discriminated in other two planes. The ground truth boundaries are the red contours. The axes are shown with blue lines.}
\label{fig:xyzplane_image}
\end{figure}

 Moreover, a novel \textit{hierarchical view-ensemble convolution} (HVEC)  module shown in Fig. \ref{fig:hvec_module} is introduced to reduce learning parameters and computation cost, and take advantage of the multi-view property  of a 3D volume. This is based on the observation that a blurred boundary on one cross-section (view)  of a 3D volume, may be easily delineated on other two cross sections, which is demonstrated in Fig. \ref{fig:xyzplane_image}. Note that the proposed strategy is completely different from the widely-used multi-view ensemble strategy \cite{huo2018splenomegaly,zhao2018deep}, which finally fuses independent 2D segmentations on three view with voting or averaging. In contrast, we focus on hierarchically extract multi-view contexts to boost the discriminative ability of the learned features. The difference between the proposed  HEVC and other group convolutions strategies \cite{gao2019res2net,szegedy2016rethinking} lies in that, we firstly decouple a 3D spatial convolution into multiple 2D spatial convolutions, and then apply different 2D convolutions on different channel groups with a focal view.
Extensive evaluations of the proposed method have been made on two public benchmarks.

The main contributions of our study can be summarized by the following threefold,

\begin{itemize}
\item We propose to use intrinsic shape cues to boost the segmentation accuracy and generalization ability,  and have  developed   a novel centerline-aware segmentation network.
\item To achieve a light-weight network for 3D segmentation, we introduce a novel HVEC module,  a simple alternative of 3D convolution to learn 3D spatial contexts using only 2D convolutions,  to reduce the number of parameters  and facilitate multi-view information aggregation.


\item Very promising performance, even with quite limited training data, in comparison  with state of the arts has
been obtained by using the proposed HIVE-Net network on two publicly available benchmarks.
\end{itemize}
This work is an extension of our preliminary work \cite{yuan2020net} with validations on more datasets and more extensive experiments on performance analysis, including the evaluation of the ability of instance-level segmentation and detection. Ablation study of our model and sensitivity analysis of tradeoff hyper-parameter are also conducted. More comprehensive review of close-related literature and detailed discussion of our motivation have also been included.

The reminder of this paper is arranged as follows. We review the related work in Section \ref{sec:relate work},  and describe the proposed methods in Section \ref{sec:method}. Dataset details, experimental results and discussions are introduced in Section \ref{sec:experiments}. Section \ref{sec:conclusion} concludes this paper.

\section{Related Work}
\label{sec:relate work}

\textbf{Methods based on hand-crafted features} mainly rely on  the extraction of highly discriminating features and a powerful classifier to automatically determine the presence of mitochondria and delineate their boundaries accurately. Beyond of using general texture features  from
computer vision studies \cite{cetina2018multi},  recent studies have introduced specially-designed geometrical and contextual features for mitochondria segmentation, such as ray features \cite{smith2009fast,lucchi2011supervoxel}, Radon-like features \cite{kumar2010radon}, local patch pattern \cite{peng2019mitochondria}, etc, which have shown greatly improved discriminative ability  \cite{peng2019mitochondria,lucchi2011supervoxel}. Given the features, shallow machine learning algorithms such as Random forest, AdaBoost,  etc,  are usually adopted for independent pixel-wise labeling.   To capture label dependencies between neighboring
pixels/voxels, conditional random fields (CRF) with manually-specified or learned parameters are usually employed. However, CRF based methods are typically computationally demanding for volume data. In order to overcome the drawbacks of shallow models and independent pixel-wise labeling,  Peng \textit{et al.} \cite{peng2019mitochondria} introduced  a cascade of structured contextual forest with novel skip connections and integrated multi-view information  to exploit contextual feature encoded in spatial-temporal predictions.
Despite  the competitive performance, hand-crafted features yet showed the limited representability, and the shallow classifiers demonstrated  limited learning capacity  for large scale data. Furthermore,  methods using iterative refinement learning \cite{peng2019mitochondria} inevitably result in high computational complexity.

\textbf{Methods based on deep representation learning} typically performs end-to-end segmentation using FCNs, and have achieved remarkable improvement in   performance. A higher performing successor  is the U-Net \cite{ronneberger2015u}, which uses skip connections to exploit multi-scale features.  Casser \textit{et al.} \cite{casser2018fast} proposed a modified 2D U-Net for mitochondria segmentation with on-the-fly data augmentation and reduced depth. Their method, however, did not take full advantage of 3D spatial context, which is crucial for medical volume segmentation. Xiao \textit{et al.} \cite{xiao2018automatic} proposed a fully residual U-Net using 3D convolution kernel and deep supervision,
which has achieved the state-of-the-art  performance. Despite  the superior performance of 3D convolutional networks, they  bring  a significant increase  of model parameters, thus need massive labeled data for good performance. For liver segmentation, Zhang \textit{et al.} \cite{zhang2019light} introduced a hybrid network that used 2D convolutions at the bottom of the encoder to decrease the complexity and 3D convolutions  in other layers to explore full spatial context. Moreover, they introduced  the depthwise and spatiotemporal separate factorization for 3D convolutions to further reduce complexity.
 Cheng \textit{et al.} \cite{cheng2017volume} introduced a light-weight 3D residual convolutional network by approximating a 3D convolution  with three 1D kernels. Modified residual connections with  the stochastic downsampling, are used to achieve feature-level data augmentation. Although using this factorized 3D kernel can greatly reduce model parameters, the 1D convolutions have  limited ability to capture crucial 3D spatial context.

 \textbf{Separable convolution} is a powerful strategy for the trade-off of light-weight architecture design and better accuracy. A standard convolution conducts computation simultaneously across the spatial dimensions and the channel dimension. \textit{Depthwise separable convolution} strategy \cite {chollet2017xception} independently performs
cross-channel convolution and  spatial convolution. In contrast,  \textit{group convolution} strategies \cite{xie2017aggregated,szegedy2016rethinking},  perform convolutions on non-overlapping groups of the  feature channels at each  layer of a network. Specifically, with channel groups, parallel filters are performed on  different filter groups. Further, in \cite{gao2019res2net}  connections are added  between different filter groups. In
aspect of spatial separation,  Inception-V3  \cite{szegedy2016rethinking} performs spatial factorization into asymmetric convolutions, e.g.  a 3$\times$3 convolution  is replaced with a 3$\times$1 convolution followed by a 1$\times$3 convolution.  In \cite{qiu2017learning}, a 3D convolution is replaced with consecutive  2D and 1D  convolutions. Instead of using the aforementioned  factorization \cite{szegedy2016rethinking,gao2019res2net,qiu2017learning}, we perform different rank-2 (2D) spatial convolutions (1$\times$3$\times$3, 3$\times$1$\times$3, 3$\times$3$\times$1) in parallel  on different  filter groups to approximate  a 3D spatial convolution as depicted in Fig. \ref{fig:hvec_module}, which captures full spatial context through  hierarchically utilizing information from three  views.  The idea of simultaneously  using 2D convolutions of different orientation was early used also in \cite{prasoon2013deep}.   A recent work \cite{gonda2018parallel} also used parallel 2D convolutions in different orientations for video
and volumetric data understanding. However, our proposed architecture is obviously different from their methods. Specifically, we perform hierarchical view-ensemble convolution with inter-connections between parallel branches and additional down-sampled branch. In this way, we can not only take advantage of  multi-range field-of-view on a  focal view, but also exploit the correlation between different views.

\section{METHOD}
\label{sec:method}

\subsection{HIVE-Net}\label{subsec:HIVE-Net}
The architecture of our proposed HIerarchical View-Ensemble Convolutional  Network (HIVE-Net) for 3D segmentation of mitochondria is shown in Fig. \ref{fig:network_img}. Our method comprises a main task for semantic 3D segmentation as well as an auxiliary centerline detection task to account for the shape information of mitochondria. We formulate the segmentation task as a voxel-wise labeling problem, and the task of centerline detection as a regression problem.

We use the skip-connected encoder-decoder architecture as (3D) U-Net \cite{cciccek20163d,ronneberger2015u}, but utilize only one shared encoder  and two task-specific decoders, where each decoder  accounts for one task. More importantly, other than using 3D convolution, we propose a novel \textit{Hierarchical View-Ensemble Convolution} (HVEC) (Fig. \ref{fig:hvec_module}) as the  building block for both the encoder and decoder, which  will be introduced in detail in Section \ref{subsec:HVEC}.

The  encoder  of the proposed HIVE-Net has $d$ (=3) down-sampling stages to extract multi-scale feature maps in gradually reduced size and capture  higher semantic information at later stages. Each decoder has $d$ (=3) up-sampling stages to gradually recover the detailed information about segmentation.  With the help of the decoders, low resolution feature maps are progressively restored to the input image size. For the down-sampling operation, we use max-pooling,  and for the up-sampling, we use trilinear interpolation. Skip-connections are used to integrate low level cues from  the encoder to the corresponding layers of decoder. Instead of using concatenation as U-Net, we use sum operation to achieve long-range residual learning \cite{he2016deep}. In fact, concatenation-based operation inevitably increases the number of feature channels, which in turn restricts the input  size.

The ground truth for  regression is  the proximity score map generated using the centerline annotations for the mitochondria, whereas the ground truth for segmentation is the voxel-wise binary labels. The proposed model is  trained with the supervision using the following loss,
\begin{equation}
L_{total} = \lambda L_{seg} + (1-\lambda) L_{reg} ,
\end{equation}
where $L_{seg}$ and $L_{reg}$  correspond to the segmentation  and  regression loss, respectively, $\lambda$ is a hyper parameter to compromise these two tasks, which is  set to 0.7 in the experiments.

\subsection{HVEC module} \label{subsec:HVEC}
Each of our proposed HVEC module  has two successive HVECs  with residual connection cross them (Fig. \ref{fig:network_img}). In each HVEC shown in  Fig. \ref{fig:hvec_module},  we firstly partition the input features into 4 groups and each group produces its own outputs, which are finally fused with concatenation. Information cross different branches are then integrated with 1$\times$1$\times$1 convolutions. Our HVEC module is different from depth-wise and group-wise separable convolutions  \cite{szegedy2016rethinking,gao2019res2net,xie2017aggregated} in four aspects: 1) instead of conducting 3D convolutions on each feature group, we perform different 2D convolutions (i.e. 1$\times$3$\times$3, 3$\times$1$\times$3, 3$\times$3$\times$1) on the first three groups to encode information of three separable orthogonal views of a 3D volume; 2) on the fourth  group of features, 1$\times$3$\times$3 convolutions are performed on down-sampled features to capture context information at large scale on a focal view; 3) to capture multi-scale contexts and multiple fields-of-view, the four branches are convoluted in serial fashion, and the feature maps convoluted by previous branch are also added to the next branch as input, resulting hierarchical connections; 4) a whole HVEC module consists of two sub-blocks, and shortcut connections across two HVEC sub-blocks are to reformulate it as learning residual function in medium range. Compared with a standard 3D convolution layer attempting to simultaneously learn filters in all the three spatial dimensions and one channel dimension, the proposed factorized scheme is more parameter efficient.
In this way, multi-scale and long-range multi-view context information, which are critical to the complicated EM image segmentation, can be encoded in a single module with significantly reduced parameters.

\subsection{Mitochondria centerline detection}
Instead of classifying each voxel as the  centerline or the background, we formulate  centerline detection as a regression problem \cite{kainz2015you}. Given the ground truth segmentation, we extract the centerline using a standard software CGAL (Computational Geometry Algorithms Library)\footnote{https://www.cgal.org/}. The ground truth for the centerline  regression is the proximity score map that is a distance transform function with peak at mitochondria centerlines and zeros on the background. Formally, it is defined as,
\begin{equation}
D(x) =
\begin{cases}
e^{\alpha(1-\frac {D_{C}(x) }{d_{M}})}-1,  & \text{if $D_{C}(x) < d_{M}$} \\
0, & \text{otherwise}
\end{cases}
\end{equation}
where $\alpha$ and ${d_{M}}$ are positive parameters to control the shape of the exponential function, and $D_{C}(x)$ represents the  minimum Euclidean distance between the voxel $x$ to the mitochondria centerline. The proximity score map $D$ is utilized as the supervision to train our centerline detection sub-network. At the end of the detection path, the final feature maps are followed by a Sigmod layer to get the predicted proximity score map $D_{out}$. Minimizing the mean squared error loss  $L_{reg}$ between $D_{out}$ and $D$ can accomplish the centerline detection task.

It is noticeable that the number of layers and channels in the detection path is smaller than that in the segmentation. Intuitively, the reasoning for this design is two fold: a) there are more smaller proportion of voxels taking  positive proximity scores than that of voxels taking positive labels. Thus, a deeper and wider network may result in over-fitting; b) increasing the number of feature channels will consume more GPU memory, and in turn decreases the size of network input, which is valuable for accurate segmentation.

\subsection{Mitochondria segmentation}
Compared with the centerline detection path, the segmentation path has one more HVEC module, as shown in Fig. \ref{fig:network_img}. Moreover, the feature maps produced by the last HVEC module in the detection path will be concatenated to the segmentation path. The goal is to take  advantage of the location and shape information contained in the detection path.
We utilize Jaccard-based loss function which is insensitive to the severe class imbalance in EM data. It is written as,
\begin{equation}\label{eq:total_loss}
 L_{seg} = 1.0 -  \frac { \sum_{i} P_i \cdot Y_i   }{\sum_{i} P_i + \sum_{i} Y_i - \sum_{i} P_i \cdot Y_i + \epsilon},
\end{equation}
where $P_i$ is the predicted probability for voxel $i$ and $Y_{i}$ is the corresponding ground truth taking binary label. The small constant  $\epsilon$ (e.g., $10^{-5}$) is to prevent dividing by zero.

\begin{figure*}[ht]
\begin{minipage}[b]{1.0\linewidth}
  \centering
  \centerline{\includegraphics[scale=0.26]{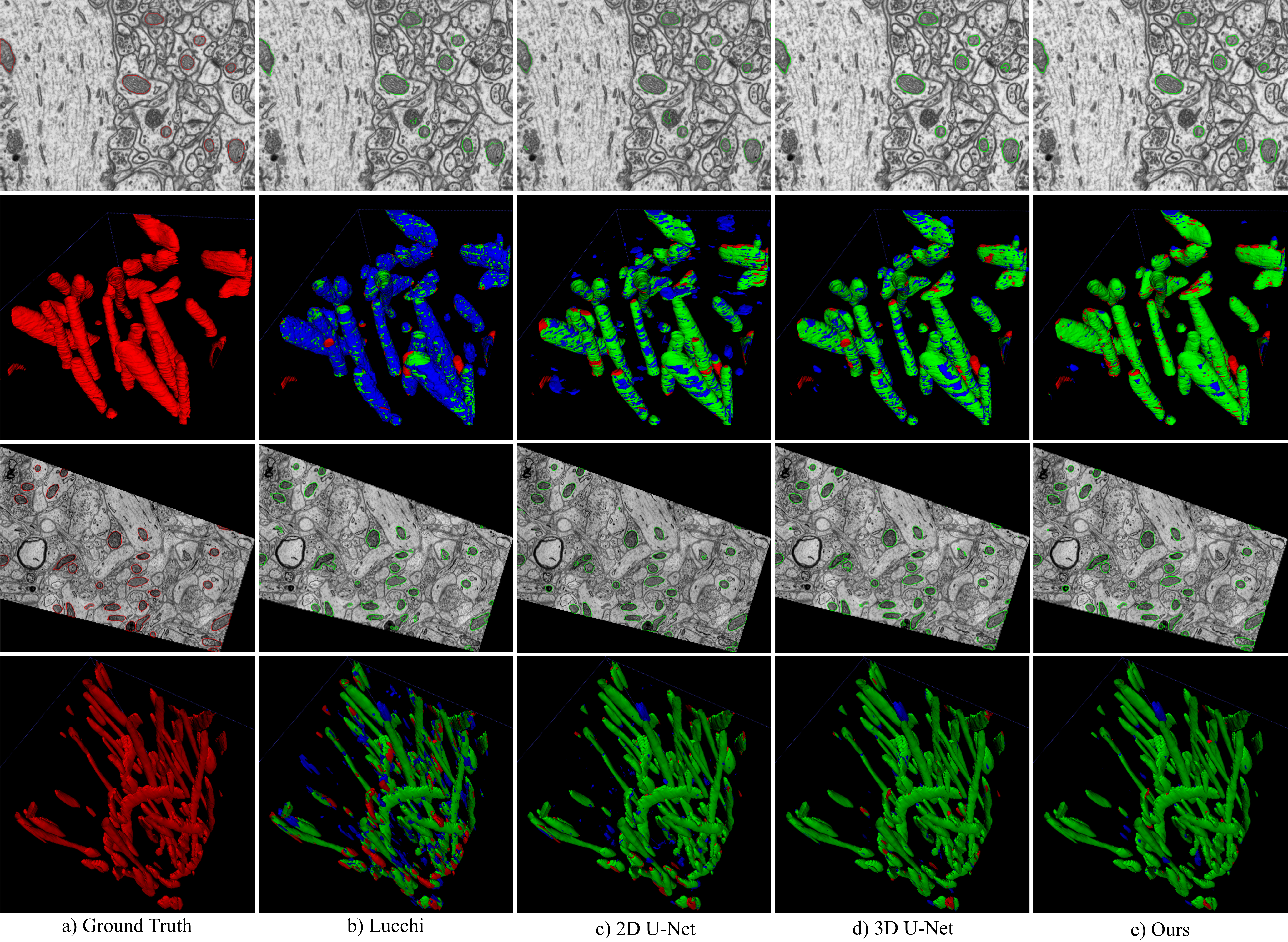}}
\end{minipage}
\caption{ The visual comparison of our segmentation result versus other methods on EPFL dataset (first and second rows) and Kasthuri++ dataset (third and last rows). In the first and third rows, the red contours denote the ground truth, and the green contours denote results of the compared methods.
The second and  fourth rows are the 3D visualization of  the  segmentations of each method and the ground truth. The red and blue fragments in columns (b)-(e) are the false negative and false positive predictions, respectively. True positive predictions are denoted by green.
}
\label{fig:segmentation_compare}
\end{figure*}

\section{Results and analysis}
\label{sec:experiments}

\subsection{Datasets}
In this section, we validate the effectiveness of our proposed methods on two public datasets, which are acquired by different electron microscopies and with different voxel spacings.

\textit{\textbf{EPFL dataset}}\footnote{https://cvlab.epfl.ch/data/em} is the most widely used  benchmark for evaluation and consists of two well-annotated stacks for model training and testing. Each stack contains 165 image slices of size 768$\times$ 1024. These images were acquired by focused ion beam scanning EM (FIB-SEM)  and  taken from CA1 hippocampus region of the mouse brain with an isotropic resolution of 5 \textit{nm}$^3$ per voxel.

\textit{\textbf{Kasthuri++ dataset}}\footnote{https://casser.io/connectomics/} was firstly released by Kasthuri \textit{et al.} \cite{kasthuri2015saturated} and the annotation of mitochondria was refined by Casser \textit{et al.} \cite{casser2018fast}. In this dataset, there are 85 image slices of size 1643 $\times$ 1613 for training and 75 slices of size  1334 $\times$ 1553 images for testing. All images were acquired by serial section EM (ssEM) and taken from 3-cylinder mouse cortex with an anisotropic resolution of 3 $\times$ 3 $\times$ 30 \textit{nm} per voxel.

\subsection{Evaluation criteria}
To measure the agreement between
the binary ground truth and automatical segmentation, two evaluation metrics are used, i.e., Dice similarity coefficient (DSC) and Jaccard-index coefficient (JAC). They are defined as,
\begin{equation}
{\rm DSC} = \frac{2 |P\cap Y| }{ |P| +|Y|}, ~~~{\rm JAC}=\frac{|P\cap Y| }{ |P \cup Y|}
\end{equation}
where $S$ and $Y$ denote the predicted binary segmentation  and ground truth, respectively. These two  metrics are widely used as assessment criteria for medical image segmentation.

To evaluate the object-level  performance of  segmentation, we have also used the aggregated Jaccard-index (AJI)  \cite{kumar2017dataset} and Panoptic Quality (PQ) \cite{kirillov2019panoptic,graham2019hover} as assessment measures. Suppose the $Y^j$ is the $j \rm th$ mitochondrion with a total of
$N$ mitochondria, the AJI is defined as follows.
\begin{equation}
{\rm AJI} = \frac{\sum_{j=1}^{N} |Y^j\cap P^{j^*}| }{\sum_{j=1}^{N} |Y^j\cup P^{j^*}| +\sum_{i\in {\rm FP}} |P^{i}|},
\end{equation}
where $j^*$ is the index of the matched segment(i.e., connected region) in the predicted segmentation $P$ with the largest overlapping (in terms of JAC)
with $Y^j$; FP is the set of false positive regions  in $P$  without the
corresponding ground truth mitochondria.

The measure of PQ  splits the
predicted and ground truth segments into three sets: true
positives (TP), false positives (FP), and false negatives
(FN ),  representing the  matched pairs of segments with at least 50\% overlapping in JAC,  unmatched
predicted segments, and unmatched ground truth segments,
respectively. Accordingly, PQ is  a combination of segmentation quality (SQ) in TP  and detection quality (DQ) in terms of F1, and is defined as follows.
\begin{equation}
\begin{matrix}
~~{\rm PQ} = \underbrace{\frac{\sum_{j\in {\rm TP}} {\rm JAC}(Y^j, P^{j^*}) }{|{\rm TP}|}}&\times&\underbrace{\frac{{\rm |TP|}}{ {\rm |TP|}+\frac{1}{2}|{\rm FP}|+\frac{1}{2}|{\rm FN}|}}\\ \scriptsize \text{~~~~~~~~~~~~~~~~~~~Segmentation Quality(SQ)}&&\scriptsize \text{Detection Quality(DQ)}
\end{matrix}
\end{equation}

To measure instance-level detection performance, we use F1, i.e., the DQ in PQ.  For completeness, we also conducted several experiments by considering different overlapping thresholds.
 Moreover, given the numbers of TP and FN, we also report the sensitivity (SEN) and specificity (SPE),
\begin{equation}
\rm SEN = \frac{|TP|}{ |TP| +|FN|}, ~~~\rm SPE=\frac{|TN| }{ |TN|+|FP|}
\end{equation}

We also use the  Average Precision (AP) \cite{lin2014microsoft}, which is a popular metric in measuring detection accuracy. The AP measures the performance with overlapping of bounding box for a detection to
be a true positive. For example, the AP-75 requires at least 75\% intersection over union (IoU) with the ground truth for a positive detection.   Moreover, we have also illustrated the result with different thresholds, i.e., AP-[50:5:85]. In comparison,  AP-65 and AP-75 are strict but proper measures, and AP-85 is too strict.
\subsection{Implementation details}
We implement our model using Pytorch \cite{paszke2019pytorch} on a workstation with 64GB RAM and one GTX 1080Ti GPU.  The model is optimized by Adam \cite{kingma2014adam} optimizer and the weight decay is set to $10^{-5}$. Initial learning rate is set to $10^{-4}$ and step-wise learning rate decay scheme is employed, where the step and decay rates are set to 15 and 0.9, respectively.
We set $\alpha$=3, and $d_M$=15.

Our network is trained using randomly cropped volumes of size
40$\times$136$\times$136 and batch size 1. Instance normalization is used.  Data augmentation including sagittal
flipping, random transpose and random rotations of 90$^\circ$ is used for model training. At the inference time, we apply test-time augmentation on   the isotropic EPFL dataset to further improve the performance, which is widely used in EM image segmentation  \cite{xiao2018automatic}, \cite{fakhry2016residual}.  For prediction efficiency, the test-time augmentation only includes  three rotations of the volume image and averaging all the predictions.


\subsection{Performance of binary and instance segmentation}
To evaluate the proposed HIVE-Net, we firstly compare it with several state-of-the-art methods on the challenging EPFL dataset, which is the \textit{de facto} standard benchmark for evaluation. Both class-level measures, i.e., DSC and JAC, and instance-level measures, i.e., AJI and PQ, are used for evaluation. These compared methods are: 1) state-of-the-art methods using  hand-crafted features including Lucchi \textit{et al.} \cite{lucchi2013learning}, Cetina \textit{et al.} \cite{cetina2018multi}, and Peng \textit{et al.} \cite{peng2019mitochondria}; 2) strong baseline FCN models including U-Net \cite{ronneberger2015u} and 3D U-Net \cite{cciccek20163d}; 3) the light-weight 3D residual convolutional network by  Cheng \textit{et al.}  \cite{cheng2017volume}  using factorized convolutions and online augmentation at
the feature-level, which is closely-related with our method, and its 2D variant; 5) the fully 3D residual U-Net by Xiao  \textit{et al.} \cite{xiao2018automatic} using deep supervision and complex test-time augmentation; 6) variants of the proposed HIVE-Net. For fair comparison, the same data augmentation at training and testing stages as our method is used in
the implementation of the compared deep learning based methods.

  \textbf{Visual comparison.} The visual comparison of our segmentation result versus three baseline methods on both datasets are shown in Fig. \ref{fig:segmentation_compare}.  As the 3D visualization in the second and  fourth rows of Fig.   \ref{fig:segmentation_compare}  shows,  the results of the 2D U-Net demonstrates a appearance with more discontinuities and false positives, due to the fact that the slice-by-slice method lacks spatial consistency between neighboring  slices and cannot consider full 3D context of a volumetric image.  In contrast, both the 3D U-Net  and our pseudo 3D method can  account for the spatial consistency between slices, enabling better results with fewer  false positives. Moreover, our method shows more accurate and more smoother results with fewer cracks, and is less memory demanding than 3D U-Net.

\begin{table}
\caption{Comparison of different methods for mitochondria segmentation on EPFL dataset. The evaluation results under both class-level measures, i.e., DSC and JAC, and instance-level measures, i.e., AJI and PQ, are reported. }
\centering
\label{tab:1}      
  \setlength{\tabcolsep}{1.2mm}
\begin{tabular}{lccccc}
\hline\noalign{\smallskip}
\multirow{2}{*}{Methods} &\multicolumn{2}{c}{Class-level}&&\multicolumn{2}{c}{Instance-level}\\
   \cline{2-3}   \cline{5-6}\noalign{\smallskip}
 & DSC($\%$) & JAC($\%$)& &AJI($\%$) & PQ($\%$) \\
\noalign{\smallskip}\hline\noalign{\smallskip}
Lucchi \textit{et al.} \cite{lucchi2013learning} & 86.0 & 75.5& &74.0&63.5\\
Cetina \textit{et al.} \cite{cetina2018multi} & 86.4 & 76.0& &--&-- \\
Peng \textit{et al.} \cite{peng2019mitochondria} & 90.9 & 83.3& &75.4 &67.7\\
U-Net \cite{ronneberger2015u} & 91.5 & 84.4&& 83.0&75.5\\
Cheng \textit{et al.} (2D) \cite{cheng2017volume} & 92.8 & 86.5& &--&--\\
3D U-Net \cite{cciccek20163d} &	93.5 &	87.8&& 86.9&80.6\\
Cheng \textit{et al.} (3D) \cite{cheng2017volume} &	94.1 &	88.9& &--&--\\
Casser  \textit{et al.}  \cite{casser2018fast}                       & 94.2 &89.0& &--&--\\
Xiao \textit{et al.} \cite{xiao2018automatic} &	94.7 &	90.0& &88.6&83.1\\
HIVE-Net (single task)	& 94.3 &	 89.2& &88.3&82.4\\
HIVE-Net (multi-task)  & 	\textbf{94.8}& \textbf{90.1}& &\textbf{89.0}&\textbf{83.9}\\
\noalign{\smallskip}\hline
\end{tabular}
\end{table}

\begin{table}[t]
\caption{Comparison of  different methods for mitochondria segmentation on Kasthuri++ dataset. The results under both class-level metrics, i.e., DSC and JAC, and instance-level metrics, i.e., AJI and PQ, are reported.}
\centering
\label{tab:2}      
  \setlength{\tabcolsep}{1.2mm}
\begin{tabular}{lccccc}
\hline\noalign{\smallskip}
\hline\noalign{\smallskip}
\multirow{2}{*}{Methods} &\multicolumn{2}{c}{Class-level}&&\multicolumn{2}{c}{Instance-level}\\
   \cline{2-3}   \cline{5-6}\noalign{\smallskip}
 & DSC($\%$) & JAC($\%$)& &AJI($\%$) & PQ($\%$) \\
\noalign{\smallskip}\hline\noalign{\smallskip}
Lucchi \textit{et al.} \cite{lucchi2013learning} & 86.2 & 75.8&&73.5&57.6\\
Peng \textit{et al.} \cite{peng2019mitochondria} & 89.3 & 80.6&&85.8 &72.9\\
U-Net \cite{ronneberger2015u} & 94.0& 88.6 && 87.5&80.2\\
3D U-Net \cite{cciccek20163d} &	94.3 &	89.2 &&87.9&81.5\\
Xiao \textit{et al.} \cite{xiao2018automatic} & 95.9 & 92.2&&91.0&85.1\\
HIVE-Net (single task)	 & 95.7&	 91.7 && 90.4&84.7\\
HIVE-Net (multi-task)  & 	\textbf{96.2}	&\textbf{92.8}&&\textbf{91.5}&\textbf{86.6}\\
\noalign{\smallskip}\hline
\end{tabular}
\end{table}
\textbf{Performance of segmentation.} The comparison results on the EPFL dataset are shown in Table \ref{tab:1}. Overall, the algorithms based on deep learning can significantly outperform traditional methods. The performance of both  the 3D U-Net and our  pseudo 3D method including its single task variant is superior than that of 2D U-Net  in all of the four metrics, which further confirms the importance of 3D spatial context and the effectiveness of the multi-view context captured by our HVEC. With additional centerline detection task, we obtain a performance gain of 0.9\% in terms of JAC and a performance gain of 0.7\% in terms of AJI and 1.5\% in PQ.  Our full model yields an accuracy of 94.8\% in DSC and 90.1\% in JAC for binary segmentation and an accuracy of 89.0\% in AJI and 83.9\% in PQ, which outperform all other methods.  These experimental results validate  the effectiveness of our HVEC module and the auxiliary detection task which can help improve the robustness and the generalization performance of segmentation task.

The comparison results on  Kasthuri++ dataset, which is a dataset with  anisotropic spatial resolution,  are summaried in Table \ref{tab:2}. On this dataset, similar results as that on EPFL dataset  have been obtained. Methods using 3D context and multi-view contexts perform better than methods relying on only 2D contexts. More specifically, the experimental results indicate that our model is superior than  other  approaches.  Our model yields an accuracy of 96.2\% in DSC and 92.8\% in JAC for binary segmentation task, and an accuracy of 91.5\% in AJI and 96.6 in PQ for object-level segmentation. With the additional centerline detection task, we obtain a performance gain of 1.1\% in JAC, 1.1\% in AJI and 1.9\% in PQ.

\textbf{Model Complexity} is typically measured by the amount of network parameters and FLOPs \cite{xie2017aggregated}. With the help of the proposed HVEC module, the parameters of our model are much less than that of  U-Net (3.2 M vs. 31 M) and 3D U-Net (3.2 M vs. 19 M). Note that we follow the settings in \cite{ronneberger2015u}, \cite{cciccek20163d},  and use four down-sampling stages in the   U-Net and three  down-sampling stages for 3D U-Net. In terms of computational complexity, our model takes about 97.7 GFLOPs, while 2D U-Net and 3D U-Net need 261.6 GFLOPs and 1244.3 GFLOPs, respectively. Although  the size (3.2 M) of our model is higher than that (1.1M) of Xiao \textit{et al.} \cite{xiao2018automatic}, their 3D model takes 134.7 GFLOPs, which is higher than our model.
Despite the significantly reduced parameters and computational complexity, our model shows state-of-the-art performance.

  \begin{table}
\caption{Detection performance on EPFL dataset. The evaluation results under DQ (i.e., F1-50), SEN, SPE, AP-65 and AP-75 are reported.  The measures DQ, SEN, and SPE are based on measuring the segment overlapping of matched instances, while the AP-65 and AP-75 are based on measuring the bounding-box overlapping of matched instances.}
\centering
\label{tab:3}      
  \setlength{\tabcolsep}{1mm}
\begin{tabular}{lcccccc}
\hline
\multirow{2}{*}{Methods} &\multicolumn{3}{c}{Segment}&&\multicolumn{2}{c}{Bounding-box}\\
   \cline{2-4}   \cline{6-7}\noalign{\smallskip}
 &DQ & SEN& SPE&&AP-65 & AP-75 \\
\hline
Lucchi \textit{et al.} \cite{lucchi2013learning} & 85.4&92.1&80.2 &&54.7&19.1\\
Peng \textit{et al.} \cite{peng2019mitochondria} & 88.5&\textbf{96.6}&82.2 &&65.5 &25.5\\
U-Net \cite{ronneberger2015u} & 88.4&93.1&84.8& & 64.1&60.2\\
3D U-Net \cite{cciccek20163d} &93.6&95.8&91.8&& 74.8&74.8\\
Xiao \textit{et al.} \cite{xiao2018automatic} &93.4&94.7&92.5&&71.3&71.3\\
HIVE-Net (single task)	& 93.4&\textbf{96.6}&90.6 &&74.9&\textbf{74.9}\\
HIVE-Net (multi-task)  &\textbf{95.0}&96.1&\textbf{94.1}& &\textbf{77.7}&\textbf{74.9}\\
\hline
\end{tabular}
\end{table}

\begin{table}
\caption{Detection performance on Kasthuri++ dataset.  The evaluation results under DQ (i.e., F1-50), SEN, SPE, AP-65 and AP-75 are reported.}
\centering
\label{tab:4}      
  \setlength{\tabcolsep}{1mm}
\begin{tabular}{lcccccc}
\hline\noalign{\smallskip}
\multirow{2}{*}{Methods} &\multicolumn{3}{c}{Segment}&&\multicolumn{2}{c}{Bounding-box}\\
   \cline{2-4}   \cline{6-7}\noalign{\smallskip}
 & DQ & SEN& SPE&&AP-65 & AP-75 \\
\noalign{\smallskip}\hline\noalign{\smallskip}
Lucchi \textit{et al.} \cite{lucchi2013learning} &68.5&79.1&60.8 &&40.7&27.1\\
Peng \textit{et al.} \cite{peng2019mitochondria} &77.9&\textbf{95.5}&66.4&&54.1&44.8\\
U-Net \cite{ronneberger2015u} &88.0&88.8&87.8&& 84.3&76.6\\
3D U-Net \cite{cciccek20163d} &90.8&92.2&89.5&& 79.9&73.2\\
Xiao \textit{et al.} \cite{xiao2018automatic} &92.3&93.3&91.5&&86.1&80.1\\
HIVE-Net (single task)	& 91.8&93.7&90.1&&84.2&80.9\\
HIVE-Net (multi-task)  &\textbf{93.3}&94.6&\textbf{92.2}& &\textbf{87.3}&\textbf{84.2}\\
\noalign{\smallskip}\hline
\end{tabular}
\end{table}
\begin{figure}[!t]
    \centering
    \includegraphics[width=0.45\textwidth]{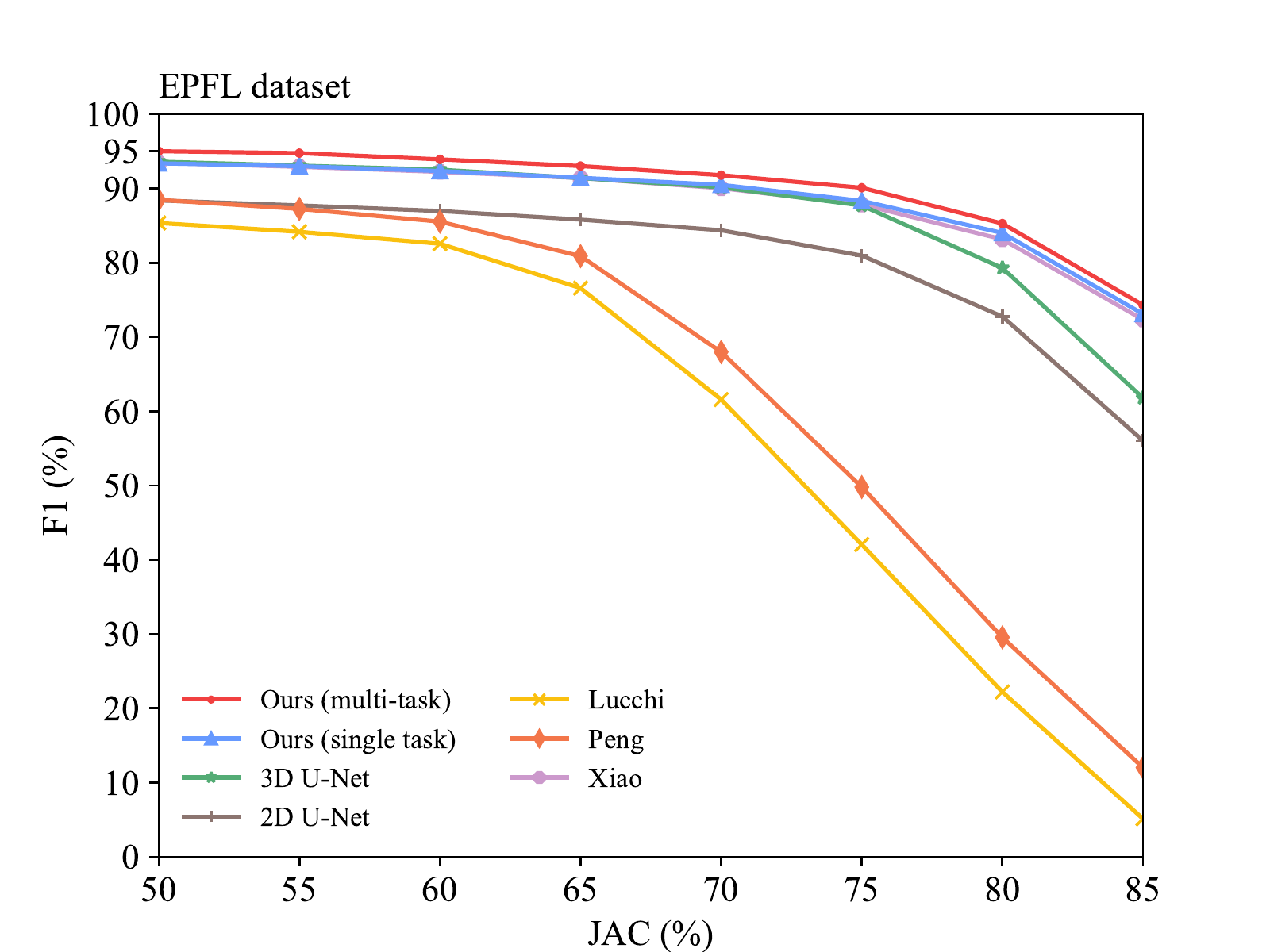}
    \includegraphics[width=0.45\textwidth]{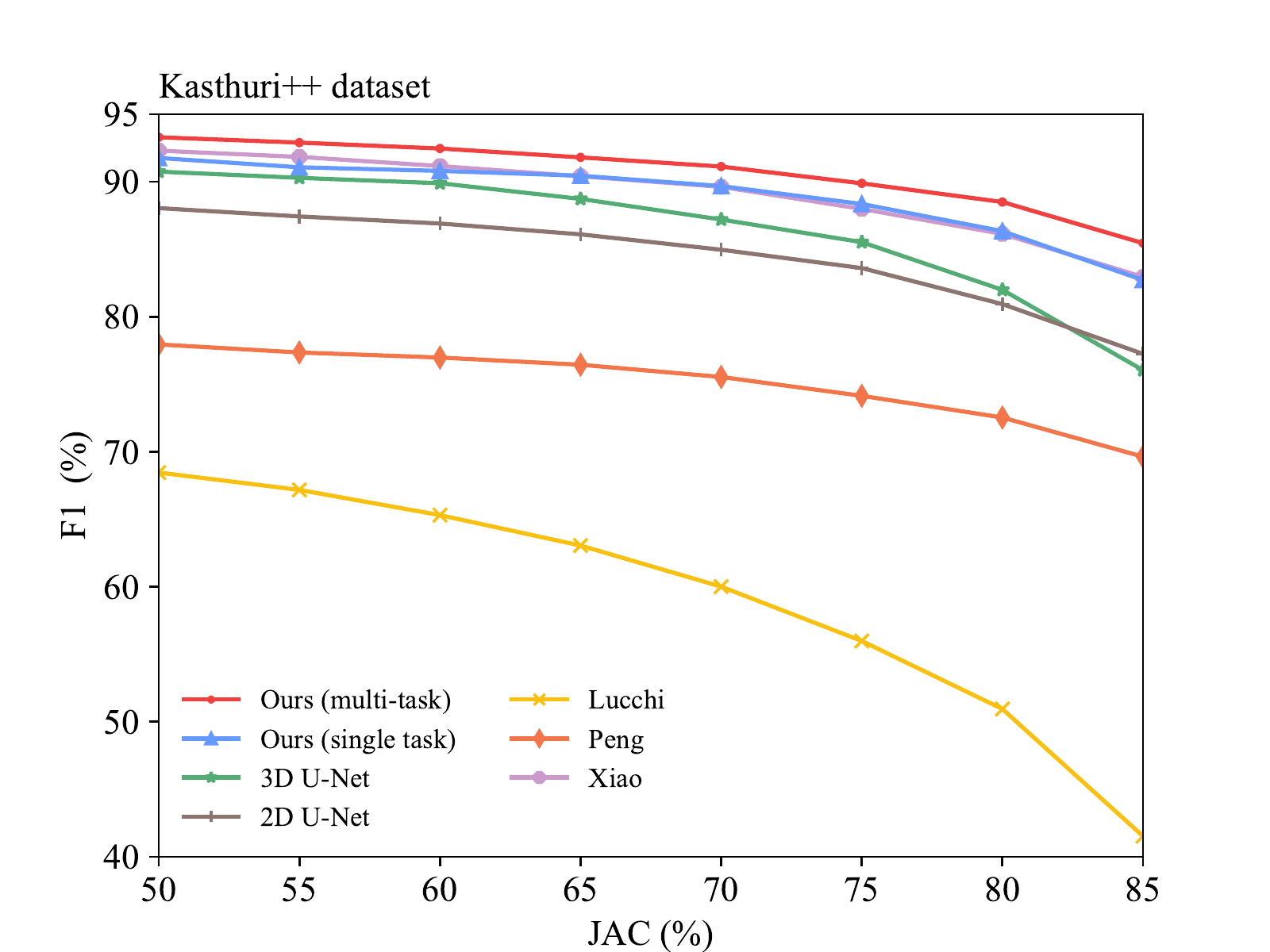}
    \caption{Detection performance in F1 with different overlapping thresholds for matched instances. The measure F1 is based on measuring the segment overlapping of matched instances. }
  \label{fig:DQ}
  \end{figure}

  Besides, we have also tested the performance of our model with reduced feature channels and thus reduced parameters. Specifically, 1) when we reduce the number of channels after the second down-sampling operation from 256 to 192 and that after the third down-sampling operation from 512 to 256, we obtain a model with 1.27 M parameters with no loss of precision; 2) when we further reduce the number of channels after the third down-sampling operation to 192, we obtain a model with 1.04 M parameters with a performance loss of 0.05\% in JAC on EPFL dataset; 3) when we further reduce the number of channels after the second down-sampling operation and that after the third down-sampling operation to 128, we obtain a model involving 0.69 M parameters and 70.1 GFLOPs but with a similar performance of 89.9\% in JAC, that is  a marginal performance loss of 0.2\% in comparison with our full model (90.1\% in JAC) on EPEL dataset. Moreover, we did not observe a loss of  performance for instance-level segmentation  in AJI (89.4\%) and PQ(84.6\%). These experimental results indicate that both our HVEC module and multi-task learning strategies are effective to improve segmentation performance.

 \subsection{Detection performance}

In addition to evaluate class-level and instance-level perfectness of the segmentation results, we also evaluate the detection performance under two class of measures. While the first class includes DQ, SEN and SPE that are based on the assessment of area overlapping of matched segments, the second class, i.e., AP, is based on the assessment of bounding-box overlapping of matched segments.

Table \ref{tab:3} and Table \ref{tab:4} have summarized the detection performance of different methods on  EPFL dataset and Kasthuri++ dataset. The DQ/F1-50 and AP of our detection on both datasets outperforms all of the compared methods, which suggests the state-of-the-art performance of our method. Specifically, on the  EPFL dataset, our model outperforms the method of Xiao \textit{et al.}  by 1.6\% in DQ/F1-50 and 3.6\% in AP-75; on the anisotropy Kasthuri++ dataset, our model outperforms the method of Xiao \textit{et al.} by 1.0\% in DQ/F1-50, and 4.1\% in AP-75.

\begin{figure}[!t]
    \centering
    \includegraphics[width=0.45\textwidth]{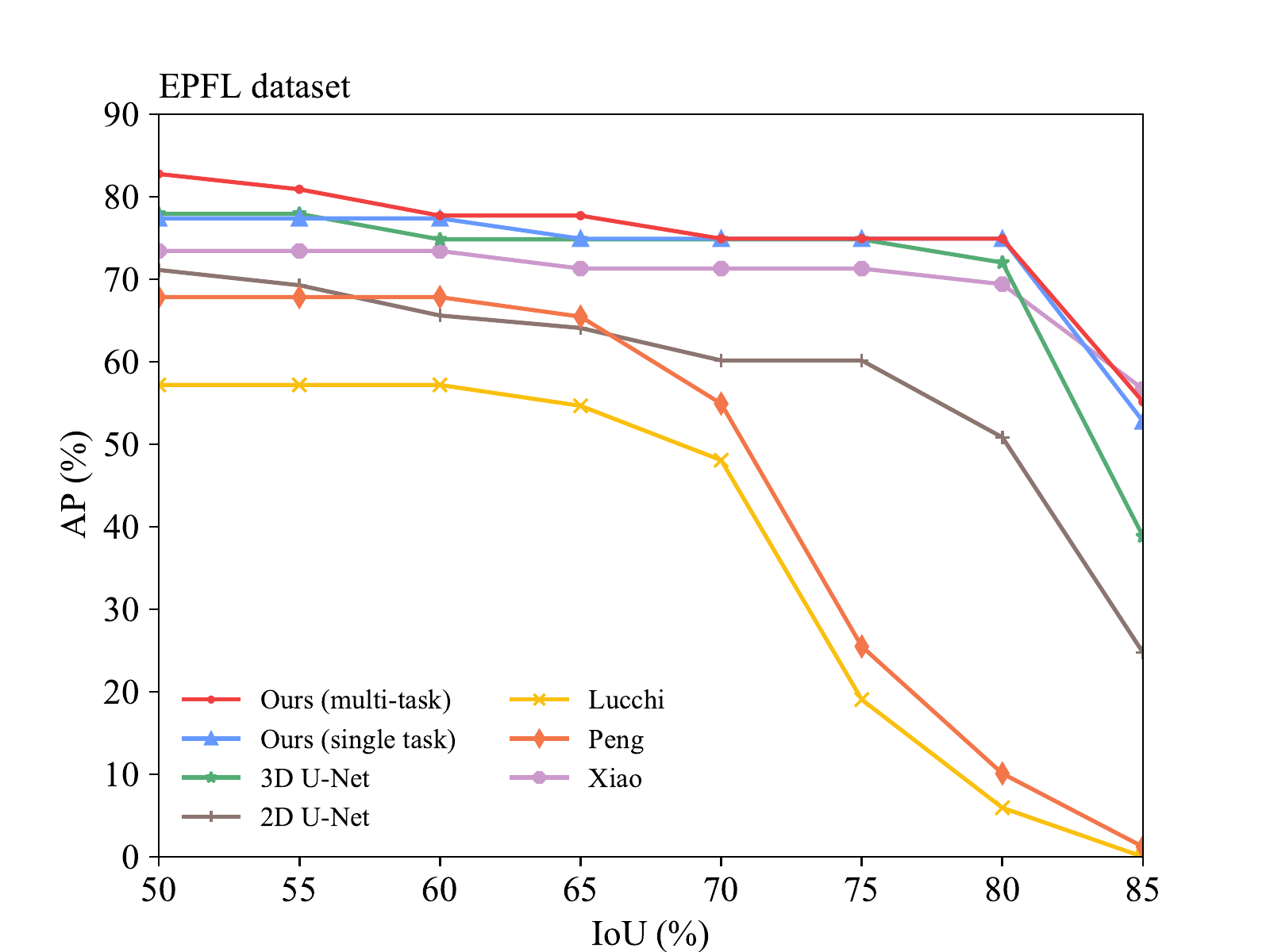}
    \includegraphics[width=0.45\textwidth]{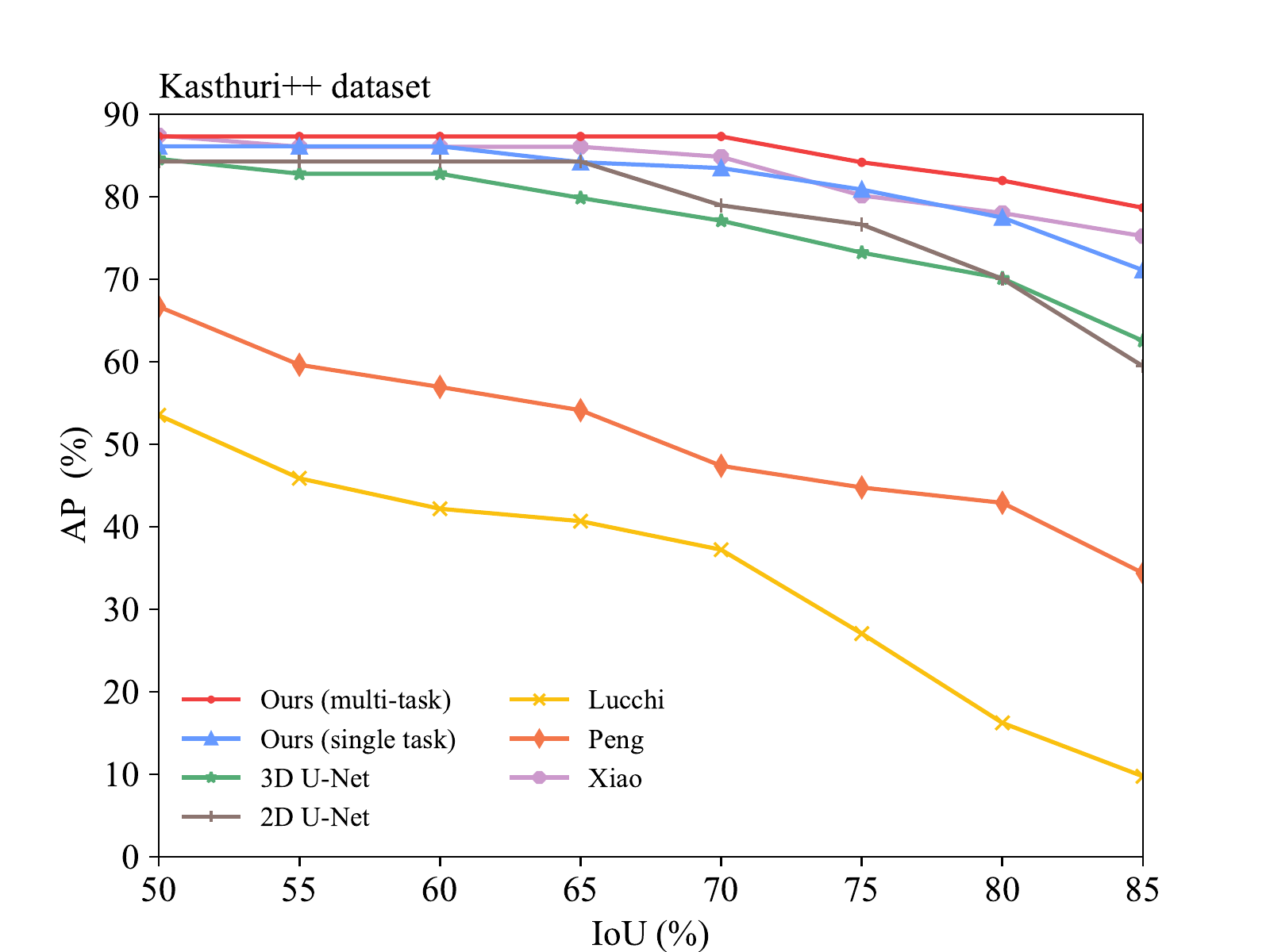}
    \caption{Detection performance in AP with different overlapping thresholds for matched instances. The AP is based on measuring the bounding-box overlapping (IoU) of matched instances. }
  \label{fig:AP}
  \end{figure}

For completeness, we also conducted several experiments by considering different overlapping thresholds ranging from 50\% to 85\% for both F1 and AP on both datasets. The results are illustrated in Fig. \ref{fig:DQ} and Fig. \ref{fig:AP}. As shown in Fig. \ref{fig:DQ}, our method yields better detection performance than the compared methods under varied thresholds. From Fig. \ref{fig:AP}, it can be seen that our method shows better results than other methods under most thresholds, especially on the more challenging Kasthuri++ dataset.

\subsection{Ablation study of the proposed HVEC block}
\begin{figure}[!t]
    \centering
    \subfigure[w/t inter-branch connections ]{
    \includegraphics[width=0.5\textwidth]{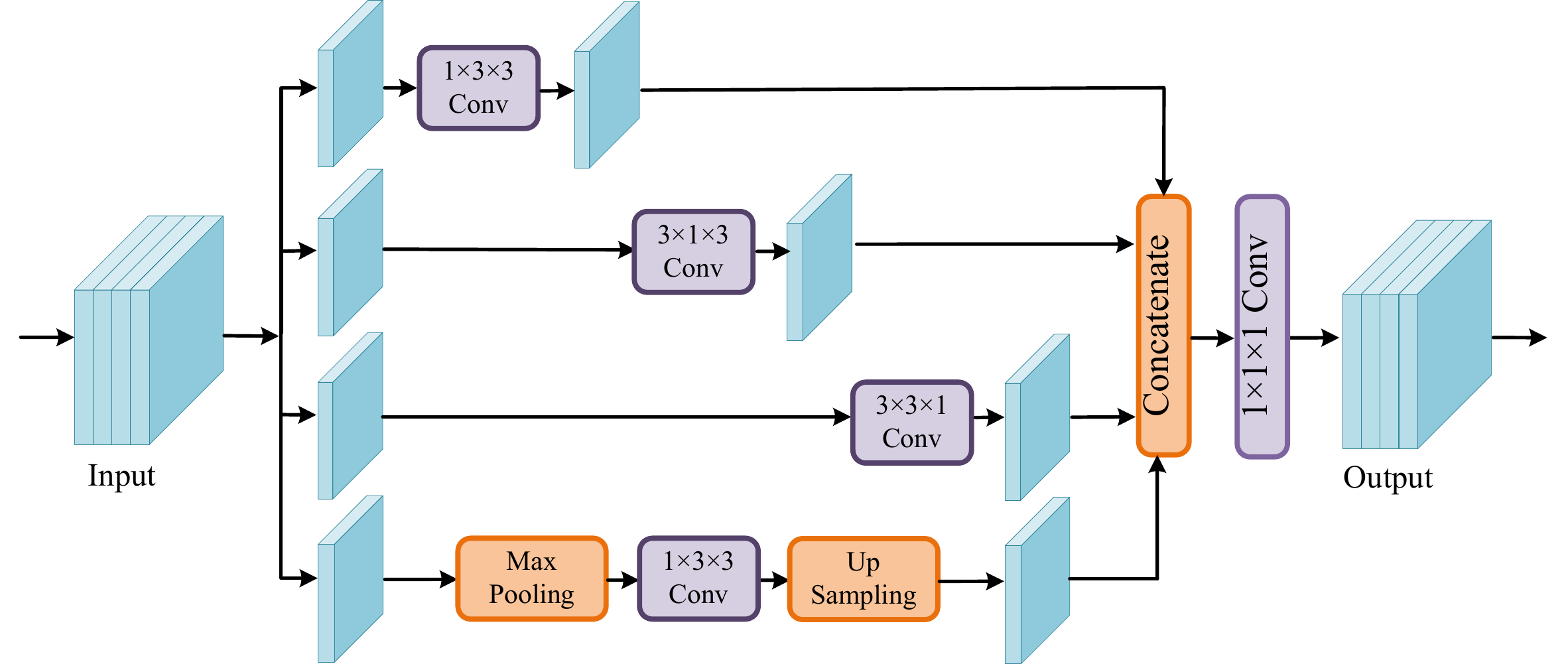}}
    \subfigure[w/t focal view branch ]{
    \includegraphics[width=0.5\textwidth]{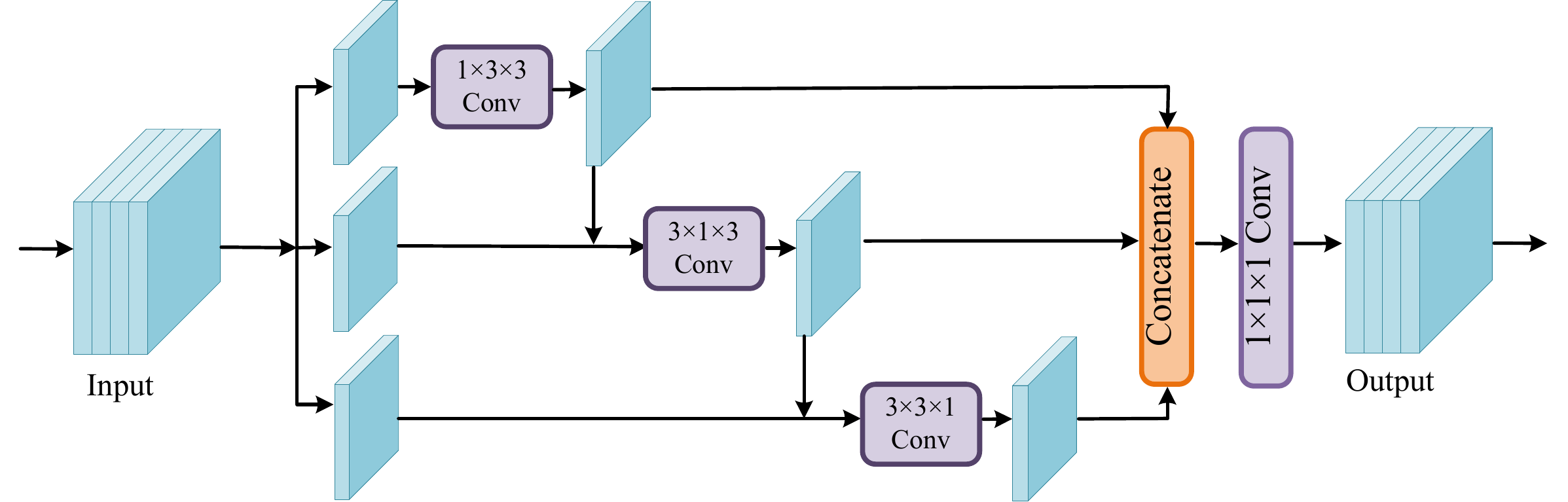}}%
    \caption{Illustration of ablated versions of the proposed HVEC block. }
  \label{fig:6}
  \end{figure}

The proposed HVEC   has a multi-branch architecture with hierarchical connections between neighboring branches and an additional branch is to extract multi-scale context on one focal view, as shown in Fig. \ref{fig:hvec_module}.  We conduct an ablation study to validate the effectiveness of two  key components, i.e.,  \textit{inter-branch connections}, and   \textit{focal view branch}, i.e.,  on the fourth branch in HVEC. Specifically, the experiments are conducted under four different settings: A) the HVEC using neither the \textit{inter-branch connections} nor the  \textit{focal view branch}; B) the HVEC without \textit{inter-branch connections} in Fig. \ref{fig:6} (a); C) the HVEC without \textit{focal view branch} in Fig. \ref{fig:6} (b); D) the full HVEC in Fig. \ref{fig:hvec_module}. The results are presented in Table \ref{ablation_hvec}.

\begin{table}[t]
    \begin{center}
    \caption{\label{ablation_hvec}Ablation study of architecture of  the proposed   HVEC block. The performance is evaluated on EPFL dataset.}
    \setlength{\tabcolsep}{1.2mm}
    \begin{tabular}{@{}l l c c c c }
    \noalign{\smallskip}\hline\noalign{\smallskip}
     &    Experiments              & A   & B & C & D \\
    \noalign{\smallskip}
\multirow{2}{*}{HVEC} &Inter-branch connections & \XSolidBrush & \XSolidBrush & \Checkmark  & \Checkmark\\
&Focal view branch & \XSolidBrush &\Checkmark &\XSolidBrush  &\Checkmark \\
\noalign{\smallskip}\hline\noalign{\smallskip}
\multirow{2}{*}{\shortstack{HIVE-Net \\(single task)}}&JAC (\%)&88.1 &88.8 &88.5   & \textbf{89.2} \\
&DSC (\%)&93.7 &94.1 &93.9   & \textbf{94.3} \\
\noalign{\smallskip}
\multirow{2}{*}{\shortstack{HIVE-Net \\(multi-task)}}&JAC (\%)&89.2 &89.8 &89.5   & \textbf{90.1} \\
&DSC (\%)&94.3 &94.6 &94.4   & \textbf{94.8} \\
\noalign{\smallskip}\hline
    \end{tabular}\\
    \end{center}
\end{table}
As mentioned in Sect. \ref{subsec:HVEC}, we use inter-branch connections in the HVEC, that is the feature maps produced by
previous branch are added to the next branch as joint input,  to enhance context feature extraction on each individual view and capture multi-scale fields-of-view. In fact, as discovered in many studies, multi-scale contexts \cite{gao2019res2net,szegedy2016rethinking,ronneberger2015u} are crucial for accurate segmentation. The effectiveness of the  hierarchical inter-branch connections  for mitochondria segmentation  is  confirmed by the performance gain, i.e., 0.4\% performance gain  in terms of JAC for our single-task model and 0.3\% performance gain for our full multi-task model (as shown in Table \ref{ablation_hvec}).

With the additional convolutional branch on a specific  view, we obtain an anisotropic model with a focal view as shown in Fig. \ref{fig:6} (b) and Fig. \ref{fig:hvec_module}.  The effectiveness of the  focal view  for mitochondria segmentation  is  confirmed by the performance gain, i.e., 0.7\%  performance improvement in JAC for our single-task model and 0.6\% performance gain for our full multi-task model (as shown in Table \ref{ablation_hvec}). By the integration of the two key components, we obtain a performance gain of 1.1\% in JAC for our single-task model and 0.9\% performance gain for our full multi-task model. All these  results show that the  hierarchical inter-branch connections  and focal view branch are critical to the performance of the proposed model.

\subsection{Sensitivity analysis of the trade-off hyper-parameter $\lambda$}

\begin{figure}[t]
  \centering
  \centerline{\includegraphics[scale=0.45]{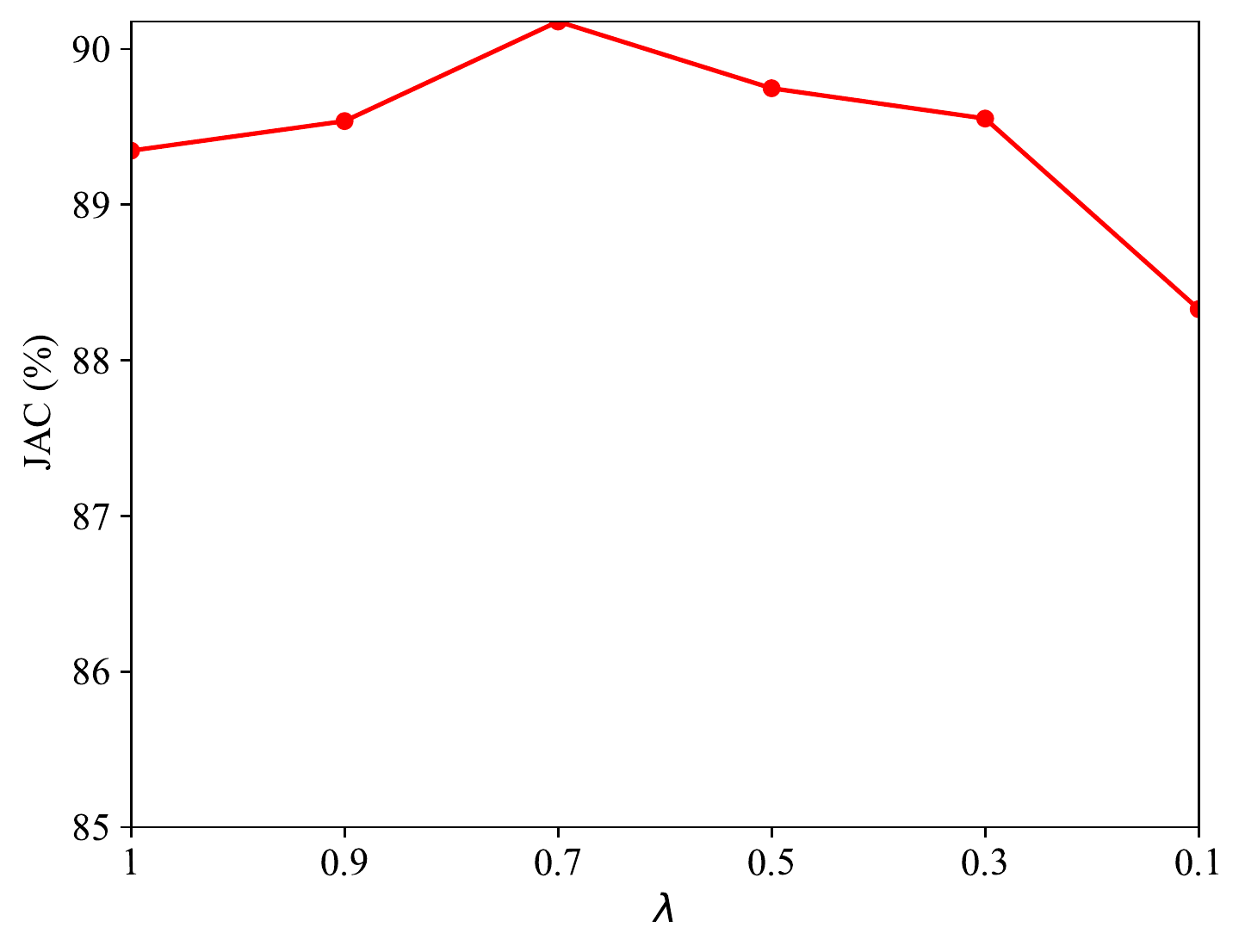}}
  \caption{Impact of the parameter $\lambda$ on the performance of our HIVE-Net.}
\label{fig:lamada_epfl}
\end{figure}

The hyper-parameter $\lambda$  in Eq. \ref{eq:total_loss}  is used to balance
the main segmentation task and the auxiliary centerline detection task. Therefore, we conduct experiments to observe the impact of changing the trade-off value of $\lambda$.
As depicted in Fig. \ref{fig:lamada_epfl}, we present the curve of JAC about $\lambda$ under different settings. Note that, when $\lambda=1$, the full model degenerates to  the single task version of our HIVE-Net.  The choice of $\lambda$ ranges from 1 to 0 , and the proposed model achieves the best with a good balance of the two tasks with $\lambda=0.7$. When $\lambda=0.1$, the centerline detection task dominates the total loss and the performance has a sharp decrease. This indicates that giving too much emphasis on the auxiliary task will reduce the segmentation accuracy  of our HIVE-Net.

\subsection{Effect of the number of training iterations}

\begin{figure}[th]
  \centering
  \centerline{\includegraphics[scale=0.45]{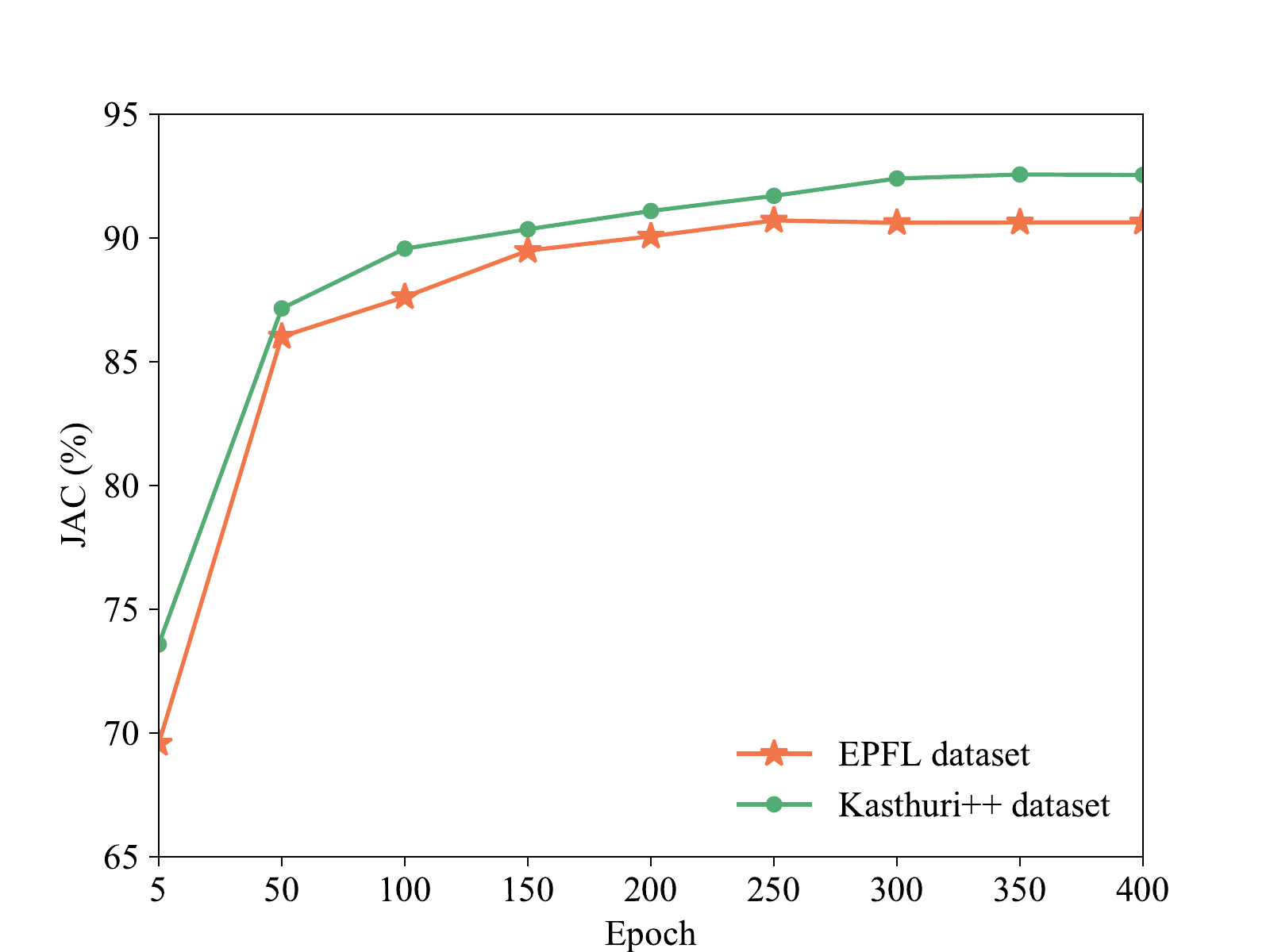}}
  \caption{Effect of the number of training iterations on the performance of our model.}
\label{fig:iteration}
\end{figure}

Since the number of training iterations usually has impact on the performance and over-training may lead to over-fitting, we conducted  multiple runs of training  with  different number of iterations  to check the stability of our results. From Fig. \ref{fig:iteration}, it is clear that with more iterations, initially there is large performance gains. However, the performance gain becomes marginal after 300 epoch. To avoid over-fitting, we use early stopping and terminate the training with a maximum of 250 epochs.

\subsection{Impact of the amount of training data}
\begin{figure}[!t]
    \centering
     \subfigure{
     \includegraphics[scale=0.45]{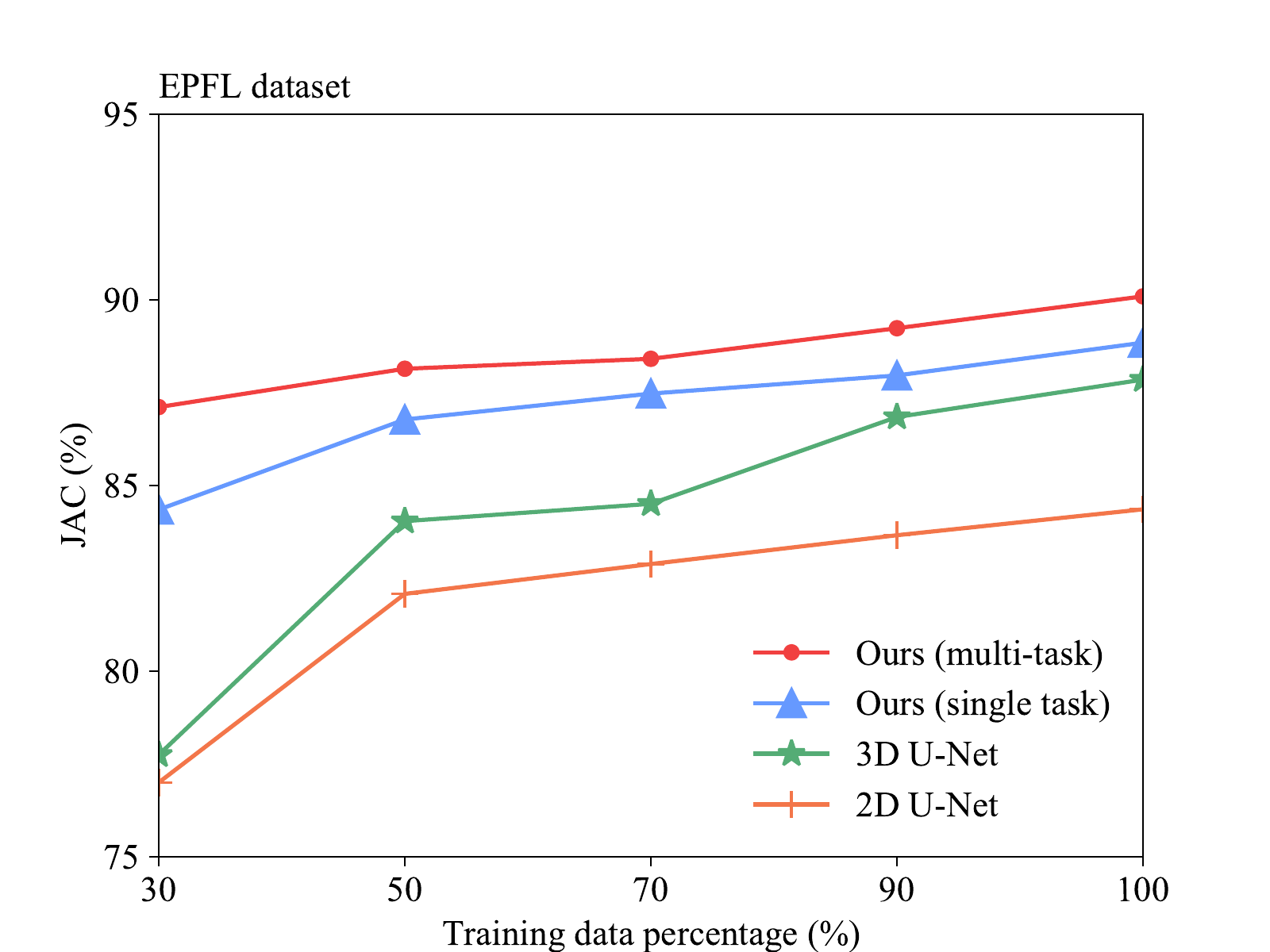}}
      \subfigure{
    \includegraphics[scale=0.45]{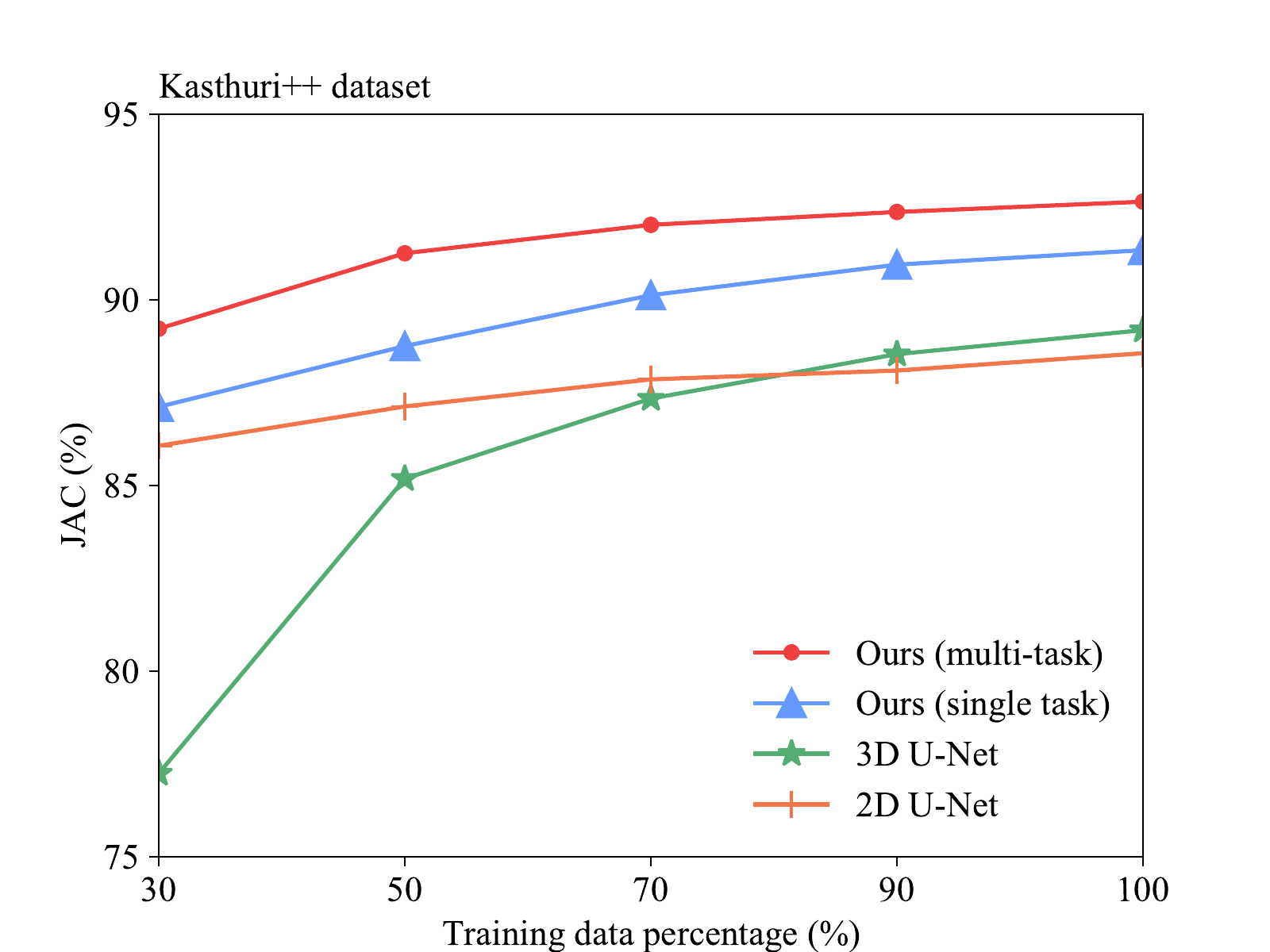}}
    \caption{The comparative mitochondria segmentation performance  on various fractions of training samples on EPFL dataset (top) and Kasthuri++ dataset  (bottom). }
  \label{fig:data_decrease}
\end{figure}

We further investigate the segmentation performance of HIVE-Net method under limited annotated data circumstances.
 Specifically, we gradually reduce the amount of training data, and test on the same testing dataset, the result of which is  depicted in Fig. \ref{fig:data_decrease}. Predictably, as the decrease of training data, all methods demonstrate  degenerated performance. Particularly, the 3D U-Net shows significant performance drop, and the gap between 3D U-Net and 2D U-Net narrows down quickly. However, our HIVE-Net and its single task variant are more robust to the reduction  of training data, and is invariably higher than both 2D U-Net and 3D U-Net. Note that, when the training data reduce to only 30\%, all of the baseline methods have a sharp drop below 80\% in accuracy. In contrast, our method still can achieve sound performance, which confirms that the proposed HIVE-Net is an effective solution for mitochondria segmentation even in cases with scarce annotated training data.

To further validate the robustness and generalization ability of our proposed model, we conduct the same experiments on Kasthuri++ dataset. With progressively decreased size of annotated training samples on Kasthuri++ dataset, as shown in Fig. \ref{fig:data_decrease}, all methods show degenerated performance. On the Kasthuri++ dataset with  anisotropic resolution, the 2D U-Net, segmenting the volume data slice by slice, shows more robust result, whereas the performance of 3D U-Net  drops dramatically   as the size of training data decrease. In contrast, even when the training data have  decreased to 30\%, our HIVE-Net still yields an accuracy over 89\% in JAC. Moreover,  our method is an end-to-end pseudo 3D model, and the input and output are both 3D image as that of the 3D U-Net.

\section{CONCLUSION}
\label{sec:conclusion}
In this paper, we have proposed an end-to-end-trainable shape-aware network for 3D segmentation of mitochondria from EM images. Besides of the main segmentation stream, our network explicitly accounts for intrinsic shape information by using a dedicated centerline detection stream. Further, the shared feature encoder between the two closely related tasks not only induces feature representations with better discriminability, but also improves the robustness and generalization ability of our model. We achieve a lightweight 3D segmentation model with fewer learnable parameters and low computation complexity by replacing 3D convolution with a novel HVEC module. Moreover, the proposed HEVC can be integrated with any 3D convolution network. Experiments show that our architecture can achieve state-of-the-art results on two challenging benchmarks, and shows significantly improved generalization ability even training with quite limited amount of training data.

A limitation of our HIVE-Net  is that it only utilizes the centerline as shape knowledge, which may have little impact on boundary fitness of the segmentation result. In the
future, the research could focus on 1) extending it to 2D networks  while maintaining the ability;  2) extend it to weakly supervised scenario, where only dot annotations are available; 3) incorporating full shape knowledge in  2D variants.

\section*{References}

\end{document}